\documentclass{article}
\usepackage[preprint]{spconf}
\usepackage{amsmath,graphicx}

\usepackage{hyperref}
\usepackage{url}
\usepackage{graphicx}
\usepackage{amssymb}
\usepackage{amsmath}
 
\usepackage{tikz,pgfplots}
\usetikzlibrary{matrix,positioning,arrows.meta,arrows,shapes.misc,spy}
\usepackage[siunitx]{circuitikz}

\usepackage{filecontents}

\usepackage{csvsimple,booktabs,array}
\usepackage{siunitx}

 \copyrightnotice{\copyright Copyright 2020 IEEE}

\title{Electric Analog Circuit Design with Hypernetworks\\ and a Differential Simulator}
\name{Michael Rotman$^1$ and Lior Wolf$^{1,2}$}
\address{$^{1}$ Tel Aviv University\quad $^{2}$ Facebook AI Research}

\begin{document}
\maketitle

\begin{abstract}
The manual design of analog circuits is a tedious task of parameter tuning that requires hours of work by human experts. In this work, we make a significant step towards a fully automatic design method that is based on deep learning. The method selects the components and their configuration, as well as their numerical parameters. By contrast, the current literature methods are limited to the parameter fitting part only.  A two-stage network is used, which first generates a chain of circuit components and then predicts their parameters. A hypernetwork scheme is used in which a weight generating network, which is conditioned on the circuit's power spectrum, produces the parameters of a primal RNN network that places the components. A differential simulator is used for refining the numerical values of the components. We show that our model provides an efficient design solution, and is superior to alternative solutions.  
\end{abstract}

\section{Introduction}
An analog circuit is an electric circuit that supports a continuous range of voltages. The information it contains is usually encoded as a time-varying signal. Analog circuits are key elements in the construction of many electronic systems. The building blocks of these circuits are electric components, such as resistors, transistors, diodes, etc. The task of designing an analog circuit is considered a difficult combinatorial task, since each component behaves differently according to the circuit configuration.

The focus of our work is the complete design of one type of an analog circuit, namely the two-port analog circuit. The two-port circuit can be depicted as a one-dimensional chain of varying linear electric components such as resistors, capacitors and inductors. While solving for the circuit's power spectrum, i.e., the output voltage and current as a function of the frequency, is a rather easy task, the inverse problem, that is, designing a circuit to hold certain properties is a challenging task. The reason is that as the number of components increases, the number of different circuit combinations rises exponentially, making brute-force approaches to circuit reconstruction unfeasible. Furthermore, even the replacement of one electric component by another in a given circuit configuration would typically result in a completely different power spectrum. 

In this work, we present an end-to-end solution to the complete design problem of two-port linear analog circuits. Instead of encoding the power spectrum into a latent space, later fed to a decoder as an input as in a traditional sequence generator, a hypernetwork $f$ encodes the power spectrum directly to the space of the recurrent neural network (RNN) decoder $g$ weights. Our contributions include
(i) unlike previous work on recurrent hypernetworks~\cite{ha2016hypernetworks}, we produce the weights of the RNN $g$ from a convolutional neural network, (ii) our method incorporates the domain knowledge using a differential simulator, which enables the interchange between discrete variables and continuous ones, and (iii) as far as we can ascertain, this is the first method to infer the circuit structure from its power spectrum.

\subsection{Related Work}
The analysis and design of analog circuits is an applied field that has been extensively studied~\cite{Razavi:2000:DAC:1594009,dactsa2012}. Circuit analysis aims to compute different induced properties, such as the output voltage and current, given a design. The inverse problem, i.e., the estimation of the components and parameters using various voltage and current measurements, has also been addressed~\cite{6403381,yong2015parameter,castejon2018automatic}. Deep reinforcement learning was previously applied to a subset of analog circuits in order to estimate circuit parameters from output measurements~\cite{learningtocircuit}. However, these methods aim to solve for the circuit's parameters, given that the circuit configuration is already known.

The hypernetwork~\cite{ha2016hypernetworks} approach utilizes one network in order to learn the weights of another network. This network design has been used successfully in many tasks in computer vision~\cite{littwin2019deep}, language modeling~\cite{suarez2017language} and sequence decoding~\cite{nachmani2019hypergraphnetwork}. While conventional sequence decoders vary the hidden state and the input sequence between recurrent steps, and the conditioning on either the initial state or the input changes between one instance to the next, hypernetworks allow for more elaborate adaption, by changing the weights of the recurrent network itself.

\section{Problem Formulation}
\begin{figure}
\begin{center}
\begin{circuitikz} \draw
(0,2) to[sinusoidal voltage source, -,l_=$V_{\text{in}}$,i_=$I_{\text{in}}$]  (0,0)
(0,2) -- (2,2)
(2,2) to[R, -,l=1<\ohm>] (4,2)
(2,2) to[C,-,l_=1<\milli\farad>] (2,0)
(4,2) to[L,-,l=0.5<\micro\henry>]  (4,0)
(0,0) -- (4,0)
(4,2) -- (4,2)
(4,2) to[short,-o,i=$I_{\text{out}}$] ++(1.5,0) node[right] {$+$}
(4,0) to[short,-o] ++(1.5,0) node[right] {$-$}
(6.0,1) node[]{$V_{out}$}
;
\end{circuitikz}
\end{center}
\vspace{-1.2em}
\caption{An example of a two-port analog circuit with three components. From left to right, an alternating power supply, $V_{\text{in}}$, a capacitator in parallel with a capacity of $1\left[ mF\right]$, a resistor in series with a resistance of $1\left[ \Omega \right]$ and an inductor with an inductance of $0.5\left[ \mu H\right]$.}
\label{fig:twoport}
\end{figure}
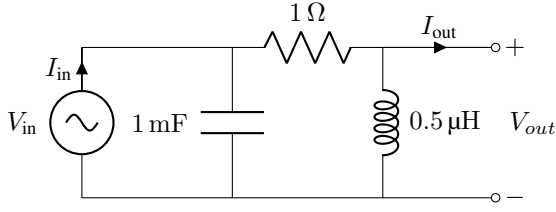

A Two-Port circuit is an analog circuit with four terminal nodes. Two are connected to an alternating power supply, and the other two are used for the circuit's output. Linear analog circuits utilize three different electric components: resistors, capacitors and inductors. Each component comes with a different numerical value (parameter), and is connected in a different alignment, either in parallel or in series. The configuration, $\mathcal{S}$, of a two-port circuit of length $n$ is an ordered list of tuples $\mathcal{S} = \left\{ {\left. {\left( {{a_i},{c_i},{v_i}} \right)} \right|1 \leq i \leq n} \right\}$. Each tuple describes the the $i$th electric component alignment($a_i$), type ($c_i$) and value ($v_i$).  An example of a circuit with a length $n=3$ composed of a capacitor, a resistor and an inductor can be seen in Fig.~\ref{fig:twoport}.

Each two-port circuit is characterized by two complex functions, $V(k): \mathbb{R}\rightarrow \mathbb{C}$ and $I(k): \mathbb{R}\rightarrow \mathbb{C}$. These functions determine the voltage and current measured over the two output nodes given a frequency $k$. The problem of designing a two-port analog circuit can be formulated as a mapping between two characteristic complex functions $V$ and $I$, as sampled at $d$ different frequencies, to the circuit configuration, $\mathcal{S}$.

\noindent{\bf Two-Port Circuit Symmetries\quad}
The number of different two-port circuits of length $n$ is $\left(2n_c n_v\right)^n$, where $n_c$ is the number of different electric components, and $n_v$ is the number of different values each of these components might yield. However, due to symmetries governed by Kirkhoff's laws, there are some circuits which are indistinguishable from one another. Given a configuration $\mathcal{S}$, any successive subset of components connected with the same alignment, in parallel or in series, could be permuted to produce $\mathcal{S}'$, which is characterized by the same voltage and current functions as $\mathcal{S}$. 

Since the required mapping is not one-to-one, we propose a canonical ordering of a circuit. Under this ordering, resistors are always followed by capacitators, which are followed by inductors. Furthermore, electric components from the same type in the same alignment configuration, are ordered by their numerical value, $v_i$. For example, the non-canonical configuration $\left\{ \left( {S,R,{1.0}} \right) \right.$ ,$\left( {P,R,0.5} \right)$, $\left( {P,C,0.1} \right)$, $\left.\left( {P,R,0.05} \right) \right\}$ contains two sub-configurations,$\left\{ \left( {S,R,{1.0}} \right) \right\}$ and  $\left\{\left( {P,R,0.5} \right) \right.$, $\left( {P,C,0.1} \right)$, $\left.\left( {P,R,0.05} \right) \right\}$, and corresponds to the canonical configuration $\left\{ \left( {S,R,{1.0}} \right. \right)$, $\left( {P,R,0.05} \right)$, $\left( {P,R,0.5} \right)$, $\left.\left( {P,C,0.1} \right) \right\}$ where $S$ and $P$ are the possible alignments (series or parallel), and $R$ ($C$) stands for a resistor (capacitator).

The number of canonical two-port circuits of length $n$ can be derived from the following generating function:
\begin{multline}
\label{eq:coefs}
    P\left( {z,{n_c},{n_v}} \right) = \frac{1}{{2{{\left( {1 - z} \right)}^{{n_v}{n_c}}} - 1}} = \\
    1 + 2{n_v}{n_c}z + \left( {{n_v}{n_c} + 3{{\left( {{n_v}{n_c}} \right)}^2}} \right){z^2} +  \ldots
\end{multline}
The number of canonical circuits of length $n$ corresponds to the coefficient of $z^n$ in Eq. \eqref{eq:coefs}. Since this is a generating function of a geometrical sum, the coefficient of $z^n$ is of the order of $O\left(n_v^n n_c^n\right)$.

\section{Method}
While small variations to the different components' numerical values $\left\{ v_i\right\}_{i=1}^n$ do not change the characteristic functions drastically, changing the component alignments and types does.
This suggests that a two-step method ought to be used: (phase I) given $V$ and $I$ sampled at $d$ frequencies, the components' alignment and type are inferred. Components' values candidates, $\left\{ \tilde v_i\right\}_{i=1}^n$, are also proposed at this stage. (phase II) to refine these values, we simulate the circuit using these candidates with a differential simulator and optimize the values only to obtain the final value $\left\{ v_i\right\}_{i=1}^n$.

Since the length of a proposed circuit configuration, $\mathcal{S}$, varies, a recurrent neural network (RNN), $g$, generates the circuit's configuration, $\mathcal{S}$. Naively, the input $x$, to such a generating network is some embedding of the characteristic functions $V$ and $I$, $f\left(V,I \right)$, so that $\mathcal{S}= g\left(f\left(V,I \right),W_g \right)$,  where $W_g$ is the set of weights of $g$. However, we suggest employing a different scheme, in which one neural network explicitly predicts the weights of another. This hypernetwork autoencoder setup essentially produces a different decoder per input, because unlike the usual autoencoder, the decoder $g$'s weights vary, based on the characteristic function $W_g = f\left(V,I \right)$.

As the weight generating network, $f$, we utilize a variant of the MultiScale Resnet(MS-Resnet)~\cite{wang2018csi}. The MS-Resnet consists of three branches, intended to capture different $1$D signal scales. Each branch is constructed using three consecutive Resnet Blocks~\cite{he2016deep}. There are three sizes of receptive fields, $3,5,7$ assigned to convolutional layers of $f$'s branches. The output of the branches is then averaged and projected to a vector space with dimension $d=256$. It is then concatenated and passed to a fully connected layer to produce an output of dimension $d=26,380$ -- the exact number of learnable parameters in the primary network, $g$.

Network $g$ is an a RNN utilizing a Gated Recurrent Unit~\cite{cho2014learning} with a hidden layer size of $64$. At each time step, the outputs of $g$ are passed to three fully connected layers, each corresponding to a different target, alignment, type and numerical value. The input of $g$ at each time step is composed of an application of a ReLU activation on the embeddings of the previous electric component.

The embedding of an electric component is the concatenation of three sub-embedding, each representing a different attribute: alignment, type and numerical value. The dimension of the alignment sub-space is $2$, whereas the dimension of the type and numerical value sub-spaces is $31$. The numerical values are quantized to five values and all three embeddings are generated by a learned look up table (LUT).

\subsection{Signal Normalization}

Since the characteristic functions image varies greatly, we normalized these functions using a $\tanh$ function, so the input to hypernetwork $f$ consists of 4 stacked channels:
\[
\tanh\left(\operatorname{Re}\left(V\right)\right) \;
\tanh\left(\operatorname{Im}\left(V\right)\right) \;
\tanh\left(\operatorname{Re}\left(I\right)\right) \;
\tanh\left(\operatorname{Im}\left(I\right)\right)
\]
 Both $V$ and $I$ were sampled on $d=512$ frequencies running on a logarithmic scale from $1\left[ \si{Hz}\right]$ to $1\left[\si{MHz}\right]$.

\subsection{Circuit Simulator}
\label{sec:circuitsimulator}
We constructed a differentiable circuit simulator using PyTorch~\cite{paszke2017automatic}.  This simulator calculates the characteristic functions $V$ and $I$ given a circuit configuration $\mathcal{S}$, which allows the estimation of the values of various components given a required signal by back-propagating throughout the simulation. Estimation of the characteristic function of a circuit can be followed by the consecutive multiplication of ABCD-parameter matrices $T$~\cite{kumar2008electric},
\begin{equation}
    \left( {\begin{array}{*{20}{c}}
  {{V_{out}}} \\
  { - {I_{out}}}
\end{array}} \right) = {T_n} \cdots {T_1}\left( {\begin{array}{*{20}{c}}
  {{V_{in}}} \\
  {{I_{in}}}
\end{array}} \right)
\end{equation}
Each of these matrices contains complex numbers, and the following representation was used:
\begin{equation}
    a + ib = \left( {\begin{array}{*{20}{c}}
  a&b \\
  { - b}&a
\end{array}} \right),
\end{equation}
i.e., replacing complex number operations with matrix ones.

\subsection{Training}
We use three cross entropy losses throughout training, where each matches a different property of the electric component,
\begin{equation}
    \mathcal{L} = \mathcal{L}_{\text{Alignment}} + \mathcal{L}_{\text{Type}} + \mathcal{L}_{\text{Value}}
\end{equation}
where each $\mathcal{L}_i$ is
$      \mathcal{L}_i = - \sum\limits_{{c_i} = 1}^{{n_i}} {{y_{o,{c_i}}}\log \left( {{p_{o,{c_i}}}} \right)}$,
and $y_{o,c_i}$ equals $1$ if and only if $c_i$ is the class of input $o$. 
Our network was trained for 700 epochs with a learning rate of $10^{-4}$ and using Teacher Forcing~\cite{williams1989learning} with a probability of $0.5$. Picking the best model over the validation set was accomplished by the model achieving the lowest partial loss function,
\begin{equation}
    \overline{\mathcal{L}} = \mathcal{L}_{\text{Alignment}} + \mathcal{L}_{\text{Type}} \,,
\end{equation}
since fixing the parameter values given the correct configuration is relatively easy.

\subsection{Inference}

While in real-life applications the numerical values of each electric component are discretized, in general the values themselves are continuous. An infinitesimal change to a component's value should result only in a slight change in the characteristic function. In order to benefit from this property, we refine the network predictions with the differentiable circuit simulator.

Given a candidate circuit configuration, $\overline{\mathcal{S}} =  \left\{({a_i},{c_i},\tilde{v}_i) \right\}$, generated by the hypernetwork decoder $g$ given $V,I$,  we optimize the following L2 loss:
\begin{equation}
  \mathcal{L}_{\mathcal{S}} = \frac{1}{d}\sum_{i=1}^d \left\lvert V_i-\mathcal{V}\left( \overline{\mathcal{S}}\right) \right\rvert^2  + \left\lvert I_i-\mathcal{I}\left( \overline{\mathcal{S}}\right)  \right\rvert^2 \,,
\end{equation}
with a Adam optimizer (learning rate of $0.01$), where $\mathcal{V}$ and $\mathcal{I}$ are the simulated characterstic functions computed by the differential simulator. Optimization stops once $\mathcal{L}_{\mathcal{S}} < 10^{-8}$.

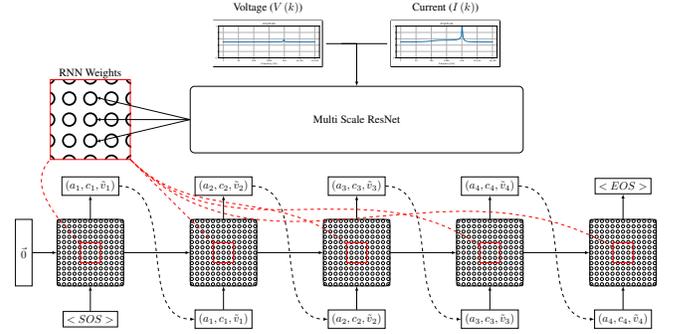
\begin{figure}[t]
\begin{center}
    
\resizebox{\columnwidth}{!}{
\begin{tikzpicture}[>=latex,y=-1cm,spy using outlines={red,  magnification=4, size=18 * 4,
                         connect spies}]
\def\bsize{60}
\def\dfromresnet{120}
\def\bdist{120}
\node[draw,rounded corners,draw,minimum width=\bsize,minimum height=\bsize,node distance=\dfromresnet] (GRU3){};
\node[draw,left of =GRU3,rounded corners,draw,minimum width=\bsize,minimum height=\bsize,node distance=\bdist ] (GRU2) {};
\node[draw,left of =GRU2,rounded corners,draw,minimum width=\bsize,minimum height=\bsize,node distance=\bdist] (GRU1) {};
\node[draw,right of =GRU3,rounded corners,draw,minimum width=\bsize,minimum height=\bsize,node distance=\bdist ] (GRU4) {};
\node[draw,right of =GRU4,rounded corners,draw,minimum width=\bsize,minimum height=\bsize,node distance=\bdist ] (GRU5) {};

\node[draw,left of =GRU1,draw,minimum width=\bsize/4,minimum height=\bsize,node distance=\bdist/2] (latent) {$\vec{0}$};

\node[below of =  GRU1,draw,node distance=\bsize] (in1){$<SOS>$};
\node[below of =  GRU2,draw,node distance=\bsize] (in2){$\left(a_1,c_1,\tilde v_1\right)$};
\node[below of =  GRU3,draw,node distance=\bsize] (in3){$\left(a_2,c_2,\tilde v_2\right)$};
\node[below of =  GRU4,draw,node distance=\bsize] (in4){$\left(a_3,c_3,\tilde v_3\right)$};
\node[below of =  GRU5,draw,node distance=\bsize] (in5){$\left(a_4,c_4,\tilde v_4\right)$};

\node[above of =  GRU1,draw,node distance=\bsize] (out1){$\left(a_1,c_1,\tilde v_1\right)$};
\node[above of =  GRU2,draw,node distance=\bsize] (out2){$\left(a_2,c_2,\tilde v_2\right)$};
\node[above of =  GRU3,draw,node distance=\bsize] (out3){$\left(a_3,c_3,\tilde v_3\right)$};
\node[above of =  GRU4,draw,node distance=\bsize] (out4){$\left(a_4,c_4,\tilde v_4\right)$};
\node[above of =  GRU5,draw,node distance=\bsize] (out5){$<EOS>$};

\draw [-latex,dashed] (out1.east) to[out=0,in=180] (in2.west);
\draw [-latex,dashed] (out2.east) to[out=0,in=180] (in3.west);
\draw [-latex,dashed] (out3.east) to[out=0,in=180] (in4.west);
\draw [-latex,dashed] (out4.east) to[out=0,in=180] (in5.west);
\draw[-latex] (GRU1.east) -- (GRU2.west);
\draw[-latex] (GRU2.east) -- (GRU3.west);
\draw[-latex] (GRU3.east) -- (GRU4.west);
\draw[-latex] (GRU4.east) -- (GRU5.west);
\draw[-latex] (latent.east) -- (GRU1.west);

\node[draw,above of=GRU3,rounded corners,draw,minimum width=5*\bsize,minimum height=\bsize,node distance=2*\bsize] (RESNET){Multi Scale ResNet};
\foreach \g in {GRU1,GRU2,GRU3,GRU4,GRU5}{
    \foreach \i in {-1.5, -1.25, ..., 1.5} {
            \foreach \j in {-1.5, -1.25, ..., 1.5} {
                \draw ({\g}.center)[yshift=\i*19,xshift=\j*19] circle (0.05);
            }
        }
}
    
\node [draw,minimum width=18,minimum height=18,very thin,red] (krembo1) at (GRU2.center)  {};
\node [draw,minimum width=18,minimum height=18,very thin,red] (krembo2) at (GRU3.center)  {};
\node [draw,minimum width=18,minimum height=18,very thin,red] (krembo3) at (GRU4.center)  {};
\node [draw,minimum width=18,minimum height=18,very thin,red] (krembo4) at (GRU5.center)  {};
\spy[spy connection path={  \draw[dashed] (tikzspyonnode) to[in=-135,out=-225] (tikzspyinnode);
                            \draw[dashed] (krembo1) to[in=-45,out=-225] (tikzspyinnode);
                            \draw[dashed] (krembo2) to[in=-45,out=-220] (tikzspyinnode);
                            \draw[dashed] (krembo3) to[in=-45,out=-215] (tikzspyinnode);
                            \draw[dashed] (krembo4) to[in=-45,out=-210] (tikzspyinnode);
                            \draw[-latex,black] (RESNET.west) -- ([xshift=5,yshift=20]tikzspyinnode.center); 
                            \draw[-latex,black] (RESNET.west) -- ([xshift=5]tikzspyinnode.center);
                            \draw[-latex,black] (RESNET.west) -- ([xshift=5,yshift=-20]tikzspyinnode.center);
                            \node[label={},black] (tttt) at ([yshift=5]tikzspyinnode.north) {RNN Weights};
                            }] on (GRU1) in node at ([yshift=120]GRU1);

\draw[-latex] (in1.north) -- (GRU1.south);
\draw[-latex] (in2.north) -- (GRU2.south);
\draw[-latex] (in3.north) -- (GRU3.south);
\draw[-latex] (in4.north) -- (GRU4.south);
\draw[-latex] (in5.north) -- (GRU5.south);

\draw[-latex] (GRU1.north) -- (out1.south);
\draw[-latex] (GRU2.north) -- (out2.south);
\draw[-latex] (GRU3.north) -- (out3.south);
\draw[-latex] (GRU4.north) -- (out4.south);
\draw[-latex] (GRU5.north) -- (out5.south);

\node[above of=RESNET,label={above:{Voltage ($V\left( k \right)$)}},yshift=40,xshift=-80] (Vcurve) {\frame{\includegraphics[angle=0,width=0.4\linewidth]{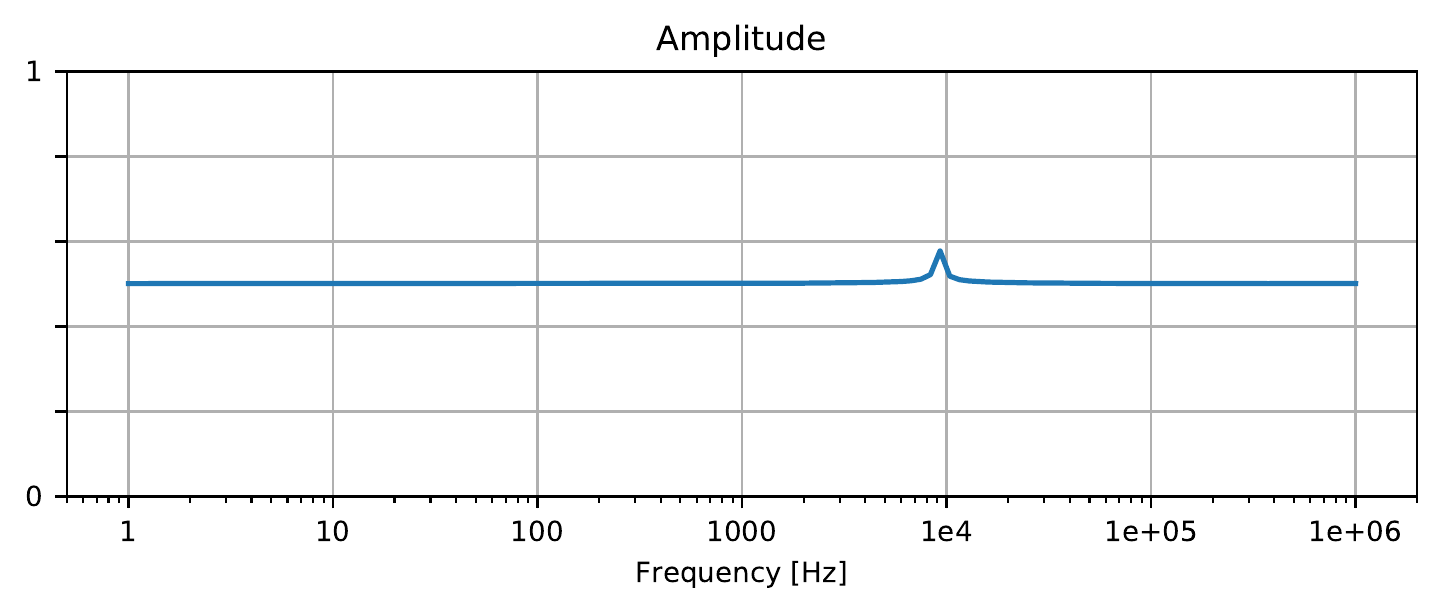}}} ;
\node[above of=RESNET,label={above:{Current ($I\left( k \right)$)}},yshift=40,xshift=80] (Icurve) {\frame{\includegraphics[angle=0,width=0.4\linewidth]{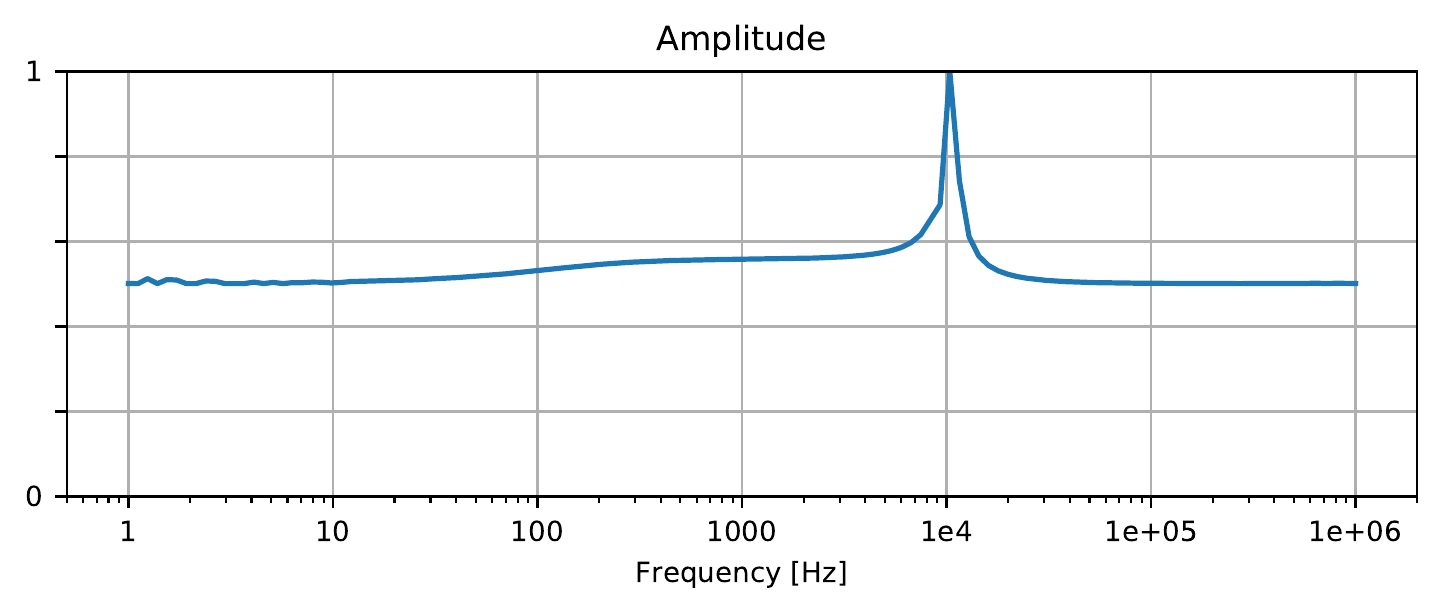}}} ;

\path (Vcurve) -- (Vcurve -| RESNET) coordinate [midway] (C);
\draw [-latex] (Vcurve.east) |- (C) -| (RESNET);
\draw  (Icurve.west) -- (C);

\end{tikzpicture}
}
\end{center}
\caption{Our proposed architecture. The voltage and current functions are sampled at $512$ frequencies and are fed into a Multi-Scale Resnet $f$, which outputs the weight matrices of a GRU $g$. The GRU outputs the circuit configuration $\mathcal{S}$.}
\label{fig:architecture}
\end{figure}
\subsection{Genetic Algorithm}
\label{sec:genetic}
A genetic algorithm was used as a baseline for comparison. The population size was set to $100$ with $10$ elite samples kept aside from each generation. The samples were mutated with a probability of $0.01$, where a mutation to the configuration could be one of the following: an addition of a random electric component, a removal of an electric component or the replacement of an electric component by another one. Breeding between different circuit configurations has also taken place at each generation. The probability of selecting a sample $x_i$ for the next generation was proportional to
$e^{-\mathcal{L}_{\mathcal{S}}}$. The algorithm was executed for 1000 generations.

\section{Experiments}

We applied our method on the canonical circuit dataset, which contains circuit configurations and their corresponding characteristic functions.
The dataset is split into three sets: training, validation and testing. The training set consists of circuit configurations of lengths $n=1,2,..,10$, where for $n=1,2,3$, all possible canonical circuit configurations were included. For $n=4,..,10$, $1,120$ random circuit configurations were drawn. In total, the training set contains $23,870$ samples. The validation set and test set contain random circuit configurations with lengths $n=4,..,10$, with $480$ and $400$ samples from each length, respectively. In total, the validation (test) set contains $3,360$ ($2800$) samples.

We evaluated our method on several scenarios. As a classical baseline, we have applied a genetic algorithm, as described in Sec.~\ref{sec:genetic}. Another baseline we experimented with is a vanilla GRU where the hidden representation obtained by $f$ is fed to the decoder $g$ as the hidden state at $t=0$. Next, we applied the hypernetwork scheme with and without our differential simulator. 
As ablation variants, we have also experimented with a variant where $f$ does not infer $g$'s classification and look up table weights, in this case, $f$ only infers the GRU weights.

\begin{table}
\caption{A comparison for the classification accuracy over the test set for the different methods.}
\label{tab:results1}
\csvloop{
file=results1.csv,
head to column names,
before reading=\centering,
tabular={@{}l@{~~}p{1.2cm}@{~~}c@{~~}p{1.0cm}@{~~}p{1.3cm}@{~~}p{1.2cm}@{~~}c@{}},
table head=\toprule Length & Genetic Algorithm &GRU& GRU per length& {Ours, GRU-only hypernet} & Ours w/o simulator & Ours  \\ \midrule,
command=\length&\ga&\vanilla & \vanillafour &\hypernonormweight& \hypernonorm& \inferfit,
table foot=\bottomrule
}
\caption{A comparison for the classification accuracy over the test set for the different methods while ignoring the classification of the numerical values of the components.}
\label{tab:results2}
\csvloop{
file=results2.csv,
head to column names,
before reading=\centering,
tabular={@{}lp{1.2cm}cp{1.2cm}p{1.3cm}c@{}},
table head=\toprule Length & Genetic Algorithm & GRU & GRU per length& {Ours, GRU-only hypernet} & Ours  \\ \midrule,
command=\length&\ga&\vanilla&\vanillafour&\hypernonormweight& \hypernonorm,
table foot=\bottomrule
}
\vspace{-.457cm}
\end{table}

As a success metric, we do not compare the characteristic functions using a distance metric since the functions vary on a logarithmic scale, and the euclidean distance between two completely different, low amplitude functions is smaller than similar high amplitude functions. In addition, even after a normalization, low frequency points contribute to the distance much more than high frequency points, which creates a highly unbalanced distance metric. Instead, we employ two different classification metrics. A complete classification is correct, when all the electric components in the configuration were correctly inferred, in their location, alignment and quantized numerical values. A value agnostic classification is correct when all the electric components in the configuration were correctly inferred, in their location and alignment. The complete classification accuracy obtained with the different methods on the test set of the canonical circuit dataset are shown in Tab.~\ref{tab:results1}. The value agnostic classification accuracy on the same set is presented in Tab.~\ref{tab:results2}.

As can be seen, our method greatly outperforms the baselines, except for length 4 where the genetic algorithm is able to cover enough of the search space. In addition, the differential render produces an additional improvement, since it manages to move some values between quantized bins. There is also clear benefit for using a hypernetwork for predicting both the set of weights of the classifier's fully-connected layers,  as well as the look up tables. This required adaptivity demonstrates that each circuit utilizes the electric component embeddings differently, and could hint that our architecture is able to learn high level ``semantics'' rather than just low-level ``syntax''. Note that one may expect that the results of our method would decay with respect to the circuits' length, however, as our model learns over a range of lengths, it best predicts at mid-range.

Despite considerable efforts, the vanilla GRU has not provided any satisfactory results. Therefore, we trained separate GRU models for different sequence lengths. As can be seen in both tables, the baseline GRUs are not competitive.

\section{Conclusions}
We proposed a method that, to the best of our knowledge, is the first method to infer the circuit's structure given a power-spectrum. Our method outperforms both evolutionary and deep learning baselines by a large margin. The differential simulator we introduce incorporates the domain knowledge in a direct way and is able to further enhance our results. The method can  be generalized in a straightforward manner to other sequence design tasks of 1D specifications.

\section*{Acknowledgments}
This project has received funding from the European Research Council (ERC) under the European Unions Horizon 2020 research and innovation programme (grant ERC CoG 725974). The contribution of the first author is part of a Ph.D. thesis research conducted at Tel Aviv University.

\bibliographystyle{IEEEbib}
\bibliography{refs}

\end{document}